# Research on Multilingual News Clustering Based on Cross-Language Word Embeddings

Lin Wu [1], Rui Li [2,*], Wong-Hing Lam [3]

1. School of Computer and Cyberspace Security, Communication University of China; wulin@cuc.edu.cn
2. School of Computer and Cyberspace Security, Communication University of China; lilrui@cuc.edu.cn
3. Department of Electrical and Electronics Engineering, University of Hong Kong; whlam@eee.hku.hk
* Correspondence: lilrui@cuc.edu.cn; Tel.: (+86)15646579159

**Abstract:** In today's world, news events are emerging incessantly. Classifying the same event reported by different countries is of significant importance for public opinion control and intelligence gathering. Due to the diverse types of news, relying solely on translators would be costly and inefficient, while depending solely on translation systems would incur considerable performance overheads in invoking translation interfaces and storing translated texts. To address this issue, we mainly focus on the clustering problem of cross-lingual news. To be specific, we use a combination of sentence vector representations of news headlines in a mixed semantic space and the topic probability distributions of news content to represent a news article. In the training of cross-lingual models, we employ knowledge distillation techniques to fit two semantic spaces into a mixed semantic space. We abandon traditional static clustering methods like K-Means and AGNES in favor of the incremental clustering algorithm Single-Pass, which we further modify to better suit cross-lingual news clustering scenarios. Our main contributions are as follows: (1) We adopt the English standard BERT as the teacher model and XLM-Roberta as the student model, training a cross-lingual model through knowledge distillation that can represent sentence-level bilingual texts in both Chinese and English. (2) We use the LDA topic model to represent news as a combination of cross-lingual vectors for headlines and topic probability distributions for content, introducing concepts such as topic similarity to address the cross-lingual issue in news content representation. (3) We adapt the Single-Pass clustering algorithm for the news context to make it more applicable. Our optimizations of Single-Pass include adjusting the distance algorithm between samples and clusters, adding cluster merging operations, and incorporating a news time parameter.

## 1. Introduction

With the rapid development of the internet, it has become the preferred method for people to publish, access, and share information. A large amount of multilingual media information contains hot topics and emotional tendencies that people are concerned about [1]. Therefore, the study of multilingual news clustering is of great significance for guiding public opinion and understanding public sentiment. News clustering algorithms can gather similar news together, preparing for subsequent tasks such as text cleaning, data mining, and data retrieval. However, when faced with cross-lingual corpora, traditional text algorithms will fail, such as TF-IDF and LDA, because they target words rather than semantics, and thus cannot recognize the relationships between cross-lingual synonyms [2]. If we use vector space models like word2vec or BERT, although they can determine a word's position in high-dimensional space through model training [3], they still cannot achieve cross-lingual embedding, and these models are often not suitable for longer texts such as news articles. Consequently, there is currently no mature, universal solution for cross-lingual text clustering problems.

This paper primarily investigates clustering solutions for news texts in both Chinese and English language contexts [4]. For short texts, pre-trained models such as word2vec or BERT construct a high-dimensional space based on semantics, with relatively accurate representation [5]. However, these models are not suitable for long texts. For clustering, a topic model can be used to cluster texts, but since topic models are based on bag-of-words models, different words with the same meaning will be treated as independent [6]. In the case of LDA, if there are English and Chinese news articles in the training corpus, Chinese topics will not be calculated for a new English news article, making topic models suitable for long texts but incapable of handling cross-lingual scenarios [7]. A new representation scheme for news domain texts is required. This paper jointly represents a news article using accurate vector representation of news headlines and topic probability distribution of news content based on topic similarity [8].

As for cross-lingual scenarios, a mixed space can be used to represent texts in multiple languages and then calculate cross-lingual similarity. The original mapping-based method assumes that synonyms in different languages have the same geometric distribution and calculates a space transformation matrix based on this assumption [9]. However, this assumption is crude, and the method was developed before BERT, with text representation based on the word2vec vector model, which is not highly accurate. Google's Multilingual BERT and Facebook's XLM pre-trained models provide a multilingual model, which, although not a cross-lingual model, can represent multiple languages [10]. In this paper, knowledge distillation techniques (teacher-student models) are used to fit Chinese and English bilingual news semantic spaces to the same semantic space, with the standard BERT model as the teacher model and XLM as the student model.

In addressing the issue of text clustering, one approach is to directly use topic models, classifying texts based on topic divisions. However, since topic models are based on bag-of-words models, they are not directly applicable in cross-lingual scenarios [11]. With cross-lingual text representation available, traditional clustering methods can be employed, such as partition-based k-means, hierarchy-based AGNES, or density-based DBSCAN. The major drawback of these clustering methods is their inapplicability to incremental clustering, while news texts are inherently real-time, with inputs consisting of individual text streams rather than static text collections [12]. If the aforementioned clustering algorithms were directly applied to incremental clustering scenarios, the cluster divisions of the original samples might change whenever a new sample point is added. This implies that all sample points must be re-clustered to ensure the correctness of the results, which would be computationally prohibitive for massive datasets and unacceptable in engineering practice [13].

In contrast, the Single-Pass clustering method is highly suitable for engineering practice, as the cluster assignments of historical samples remain unchanged. Consequently, when clustering new samples, the time complexity is merely $O(n)$ (n being the current number of samples). Building on the Single-Pass approach, this paper improves its time complexity to $O(k)$ (k being the current number of clusters) and incorporates a temporal dimension for news articles, enabling more precise fine-grained clustering at the level of news events.

## 2. Related Works

In this section, we briefly review the previous methods most relevant to our work, including text representation, cross-language word embeddings, and text clustering.

*2.1. Text representation*

To perform text clustering, the first issue to be addressed is text representation [14]. In the 1970s, Salton and others proposed the vector space model, transforming text into a point in high-dimensional space, and turning text computations into vector calculations. This idea has evolved into two specific approaches to date [15].

One approach is the traditional VSM (Vector Space Model). This involves tokenizing the document, with each word constituting a dimension in the space. Feature selection is then carried out based on document frequency and information gain, among other criteria, ultimately generating a vector to represent the text [16]. Although the VSM model is simple, it has a notable drawback. For instance, consider the phrases "I like playing basketball" and "He loves running." Both sentences indicate someone's preference for a particular sport, but they do not share any common words. In the VSM model, their feature vectors are orthogonal and have no similarity. Therefore, using this approach to represent text is not advisable in multilingual scenarios [17].

Another approach is to represent words as relationships between adjacent words. In 2013, Mikolov and others introduced word2vec, predicting one word from another and determining the meaning of a word from its co-occurrence relationships [18]. This led to the development of two word2vec algorithms: CBOW (Continuous Bag of Words) and skip-gram. CBOW predicts the center word from its surrounding words, while skip-gram predicts the surrounding words from the center word. The emergence of word2vec compensates for the shortcomings of the traditional VSM model.

*2.2. Cross language word embedding*

The fundamental task of cross-lingual word embeddings is to construct a unified vector space in which synonymous words from different languages have the same vector representation [19]. There are two basic approaches to solving this problem: mapping-based methods and pseudo-multilingual corpus-based methods.

Mapping-based methods involve training individual monolingual vector spaces and then mapping them together using a certain technique [20]. In 2013, Mikolov and others discovered that the geometric arrangement of English and Spanish words in their respective vector spaces is strikingly similar, allowing for one space to be transformed into another through a linear mapping. The transformation matrix is obtained by minimizing the mean squared error of bilingual representations using gradient descent, which relies on bilingual dictionaries. To reduce dependency on bilingual vocabulary, Vulic and Korhonen combined BWE models and used document-level aligned corpora, eliminating the need for seed vocabulary. The advent of BERT (Bidirectional Encoder Representations from Transformers) introduced a new pre-training and fine-tuning paradigm. Building upon BERT, Google released a multilingual BERT model suitable for 109 languages, and Facebook released an improved XML-R, both providing embeddings for more than 100 languages. In 2020, Reimers and others aligned cross-lingual vector spaces using parallel sentence pairs based on multilingual pre-trained models, and through knowledge distillation, the student model learned the precise monolingual representations from the teacher model [21].

The other approach involves constructing a pseudo-multilingual corpus by mixing multilingual data as new training data, and then directly training a vector space capable of representing multiple languages. Gouws and others first applied this method to part-of-speech tagging tasks, constructing a dictionary to describe word equivalence classes for generating mixed context-target pairs [22]. For each word w, its equivalence class is randomly replaced, and then the CBOW algorithm from word2vec is used to train the mixed corpus. Ammar and others trained a multilingual model by clustering synonyms from different languages into the same cluster, then using skip-gram to train the cluster identifiers, resulting in similar vector representations for synonymous words from different languages.

One approach focuses on fitting the embedding space for each language, while the other focuses on the dataset directly. Due to the complexity of constructing pseudo-multilingual datasets, their quality is often low, resulting in less accurate cross-lingual embedding spaces. With the emergence of various pre-trained models providing high-quality embeddings, the need for training monolingual models is reduced. In recent years, the research community has favored approaches that fit multilingual embedding spaces based on monolingual models [23].

*2.3. Text clustering*

Text clustering can be broadly divided into two types of algorithms. One type relies on the vector representation of the text, including traditional partitioning, hierarchical, and density-based clustering algorithms [24]. These algorithms treat the objects to be clustered as independent points in an n-dimensional space, using Euclidean distance, cosine similarity, and other metrics to define the distance between objects. This approach requires vectorizing the text first, and the accuracy of the text representation directly affects the clustering results. Representing long texts as vectors can be challenging [25]. The second type of algorithm is model-based, such as the LDA topic model. This approach assumes that a text is composed of a probability distribution of a series of topics, without requiring vector representation of the text and without limiting text length. However, this approach is still based on word computation, and during the training process of the topic model, it cannot recognize synonymous words in different languages, such as Chinese and English. Consequently, if the training corpus contains multiple languages, even if semantically they belong to the same topic, LDA will still allocate them to different topics [26].

**3. Methods**

In this section, we detail the representation of news, the similarity calculation of cross-language news, and the clustering of Chinese-English bilingual news based on the improved Single-Pass.

*3.1. Representation of news*

3.1.1. Representation of news headline

News headline is a generalization of news and the text length is short, which conforms to the sentence vector limit input length (128 characters), so this paper uses the sentence vector of the cross-language model to represent a news headline [27].

Cross-lingual word embeddings are implemented using knowledge distillation on pretrained model. The teacher model is bert-base-nli-stsb-mean-tokens. This pre-training model is based on the standard BERT, using Sentence BERT technology for fine-tuning to make BERT's sentence vector representation more suitable for semantic similarity tasks. The training data of tune is the public data set of nli and stsb, and the sentence is encoded by the token mean method. The student model is XLM-Roberta-base [28].

The goal of this paper is to let the Chinese and English representations of the student model XLM-Roberta learn the monolingual accurate representation of the teacher model through knowledge distillation [29]. The data set of the model is about 100,000 Chinese-English translation news headline pairs, the test set is about 20,000 Chinese-English translation headline pairs, the Chinese title is $\{c_1, c_2, ......c_n\}$, the English title is $\{e_1, e_2, ......e_n\}$, the title x is input into the teacher model, and the sentence vector is written as. The sentence vector input into the student model is written as, and the following loss function is obtained according to knowledge distillation:

$$Loss = \sum_{i=1}^{n}\left(\left\|S_{c_i} - T_{e_i}\right\|^2 + \left\|S_{e_i} = T_{e_i}\right\|^2\right) \quad (1)$$

There is a pooling layer after the input layer, and the pooling layer uses the token average scheme to express the sentence vector. The structure of the model is shown in Figure 1 below:

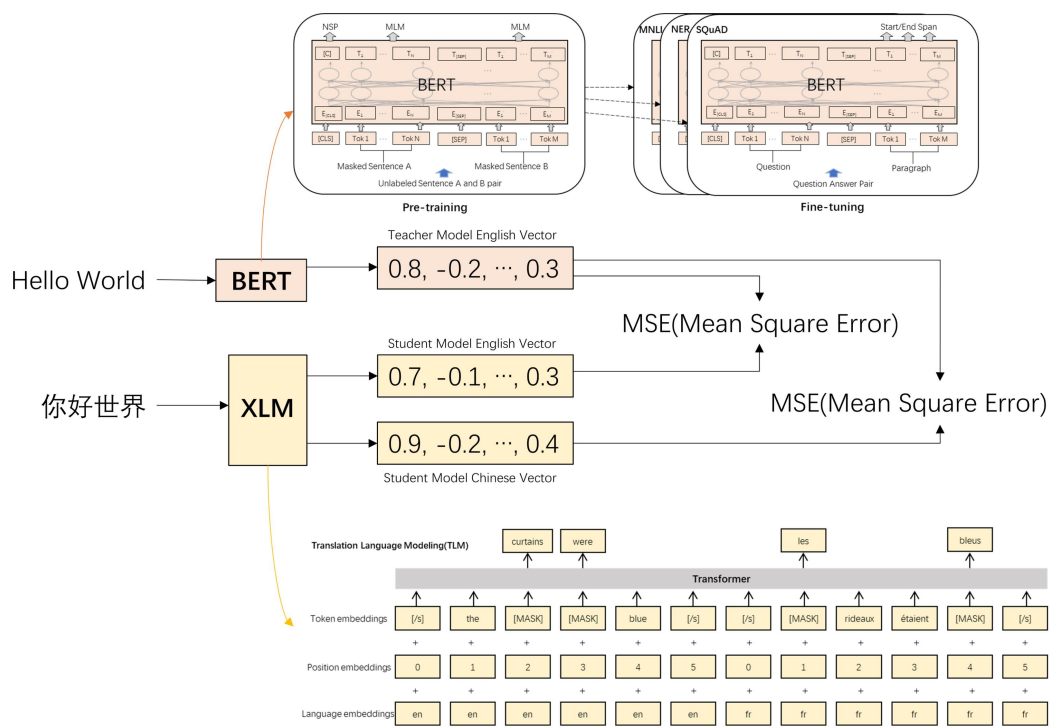

**Figure 1.** The structure of the model.

In order to verify the validity of the model, this paper crawled 1021 headlines from People's Daily Online from January 10 to January 31, 2022, and translated them into English using Baidu Translate API. The similarity is used to match sentence pairs. The calculation of text similarity uses the cosine similarity of sentence vectors. The matching accuracy of sentence pairs reaches 96.67%.

3.1.2. Representation of news content

For the content of news, this paper uses LDA topic model to express. LDA can be divided into two processes: training and inference [30]. Training finally obtains a topic-word probability distribution matrix and a document-topic probability distribution matrix, that is, each topic is composed of the probability distribution of words, and each document is composed of the topic's probability distribution matrix [31]. Probability distribution composition. Since the document-topic distribution matrix has no effect on inferring the topic of a new document, only the topic-word distribution matrix is saved; during the inference process, Gibbs sampling is performed on the word to infer its topic distribution [32].

The source of the training corpus is 40,013 news articles on seven topics of economy, politics, military, entertainment, sports, science and technology, and automobiles published by seven domestic and foreign news media. In order to ensure the strict symmetry of Chinese and English materials, Baidu Translate API is used to translate Chinese corpus into English, English corpus into Chinese, and HanLP is used for word segmentation to train the LDA model. After experimental comparison, it is found that it is most appropriate to set the number of topics K as 22. Ideally, the number of topics should be 14, but because the topic of a tag itself is incomplete, such as political news, the difference between the word-document count matrix of domestic and international political news is very large, and for example, the word-document count matrix of news that is biased towards art and idol-chasing in entertainment is also very different, and the word-document count matrix is the only real observation variable in the entire LDA. This shows that the news under the same label on the same website is also different. If we set the number of topics K as 14, there will be problems that the topic under a certain label disappears and two themes appear under a certain label.

The test set of the LDA model is 2100 news articles evenly distributed under the above seven topic labels, of which 1050 are used for the test of Chinese topics, and the other 1050 are used for the prediction of English topics. The subject of the news is compared with the real label, and the accuracy of the classification is 91.57%.

Different from binary classification, since there are not only positive and negative samples in multi-classification problems, indicators such as recall rate and F1 value cannot be directly applied to the evaluation of multi-classification models [33]. One solution is to convert the multi-classification problem into a binary classification problem for evaluation, such as precision (average of various types of accuracy), recall (average of various recall rates), macroF1 value (average of various types of F1 values) Another The scheme directly defines multi-class indicators, such as Kappa coefficient:

$$k = \frac{p_0 - p_e}{1 - p_e} \tag{2}$$

where is the classification accuracy for all samples, and is the sum of the corresponding row and column dot products in the confusion matrix.

The experimental results are shown in Table 1 below:

**Table 1.** The experimental results.

| Real/Forecast | Economy | Politics | Military | Entertainment | Sport | Science | Automobile |
|---|---|---|---|---|---|---|---|
| Economy | 281 | 9 | 2 | 0 | 0 | 5 | 3 |
| Politics | 3 | 273 | 12 | 0 | 2 | 9 | 1 |
| Military | 1 | 4 | 288 | 1 | 1 | 1 | 4 |
| Entertainment | 2 | 0 | 0 | 291 | 5 | 2 | 0 |
| Sports | 7 | 2 | 6 | 7 | 258 | 16 | 4 |
| Science | 4 | 11 | 6 | 0 | 1 | 269 | 9 |
| Automobile | 3 | 0 | 1 | 1 | 7 | 25 | 263 |

After calculation, $p_0 = 0.9157, p_e = 0.1429$, Kappa coefficient $k = (0.9157 - 0.1429)/(1 - 0.1429) \approx 0.9016$.

To sum up: for any piece of news N, its title T can represent a vector in the semantic space of the cross-language model:

$$T = [t_1, t_2, \ldots t_n] \tag{3}$$

Its content C can be expressed as the topic probability distribution of the LDA topic model:

$$C = [c_1, c_2, \ldots c_k] \tag{4}$$

where n is the dimension of the semantic space and k is the number of topics specified by LDA.

*3.2. Similarity calculation of cross-language news*

3.2.1. Topic similarity calculation

LDA model is a bag-of-words model. Words are independent of each other and cannot solve cross-language problems [34]. In this paper, by setting the training corpus as a bilingual parallel corpus, there are no identical words between bilingual documents, so the topics of the bilingual texts are also independent, so that the bilingual topics present a symmetrical distribution. This process is equivalent to using a symmetrical bilingual corpus to train two If there are two LDA models, the topics in one model are all Chinese, and the topics in the other model are all English.

Suppose the number of topics is 2k, the Chinese topic sequence $Topic_{cn} = \{cn_1, cn_2, \ldots cn_k\}$, the English topic sequence is $Topic_{en} = \{en_1, en_2, \ldots en_k\}$, where $en_i$ is the symmetric topic of $cn_i$.

The problem can now be transformed into a bilingual topic sequence $Topic = \{unknown_1, unknown_2, \ldots unknown_{2k}\}$, the sequence is transformed into $\{t_1, t_2, \ldots t_k, t_{k+1}, t_{2k}\}$, so that $t_i$ and $t_{i+k}$ are bilingual symmetric topics. Suppose that there is a matrix $TopicMatrix_{2k \times 2k}$ so that $TopicMatrix_{ij}$ is the similarity between topic i and topic j, then this matrix is called topic similarity matrix. For any topic i, we can find the most similar topic, thus solving this problem.

Topic similarity can be defined by the similarity of subject words. In LDA, a topic is embodied as a certain probability distribution of words, sorted by probability from large to small, and the top m words with the largest frequency are defined as the topic words of a topic [35]. A subject can thus be defined as a collection of subject terms:

$$topic = \{tw_1, tw_2, \ldots, tw_m\} \quad (5)$$

Subject headings can be directly represented by a cross language model, and the process of representing a vector of text x through a cross language model (Cross Language Model) is defined as the following function:

$$CLM(x) = (a_1, a_2, \ldots a_{768}) \quad (6)$$

The similarity of subject words is defined by cosine similarity, which uses the angle between two vectors in a high-dimensional space to describe how similar two vectors are:

$$cossim(A, B) = cos(\theta) = \frac{\sum_{i=1}^{n} A_i \times B_i}{\sqrt{\sum_{i=1}^{n} (A_i)^2} \times \sqrt{\sum_{i=1}^{n} (B_i)^2}} \quad (7)$$

The similarity between subject words $a_i$ and $b = (b_1, b_2 \ldots b_m)$ is defined as the similarity between subject words $a_i$ and b:

$$cossim\_word2topic(a\_i, b) = max(cossim(a\_i, b\_1), cossim(a\_i, b\_2), \ldots, cossim(a\_i, b\_m)) \quad (8)$$

Then the similarity between topic a and topic b is:

$$cossim\_topic2topic(a, b) = \frac{\sum_{i=1}^{m} cossim\_word2topic(a_i, b)}{m} \quad (9)$$

To guarantee $cossim\_topic2topic(a, b) = cossim\_topic2topic(b, a)$, the topic similarity is defined as:

$$topicsim(a, b) = \frac{\sum_{i=1}^{m} \left(cossim\_word2topic(a_i, b) + cossim\_word2topic(a, b_i)\right)}{2m} \quad (10)$$

In order to make the distribution of topic similarity between 0 and 1, and the similarity between asymmetric topics is lower and the similarity between symmetric topics is higher, the sine function is used to remap the topic similarity [36]. Calculate the similarity between all topics, the minimum value is $sim_{min}$, the maximum value is $sim_{max}$, the similarity is evenly distributed between $-\pi/2$ and $\pi/2$, through the following linear function $y = ax + b$, where $a = (\pi)/(sim_{max} - sim_{min})$, $b = 0.5\pi - sim_{max} \times a$. Use the sine function to map and normalize the data between 0 and 1:

$$y = 0.5sin(ax + b) + 0.5 \tag{11}$$

so there are:

$$topicsim(a,b) \coloneqq \frac{1}{2}sin\left(\frac{topicsim(a,b) - sim_{max}}{sim_{max} - sim_{min}} + \frac{\pi}{2}\right) + \frac{1}{2} \tag{12}$$

This paper does not map the distribution of similarity to 01 distribution, because there is also a certain similarity between different topics, such as technology topics and automobile topics, military topics and political topics.

1. For each topic of LDA, take the top m words with the highest probability as the topic words of the topic;
2. For any topic word w, it can be Through the cross-language model, it is represented by a 768-dimensional vector;
3. Calculate the similarity between all topics;
4. Count the maximum similarity and minimum similarity in all samples, and re-normalize the similarity to Between 0 and 1.

3.2.2. News similarity calculation

News can be expressed as the vector representation of the semantic space of the title $T = [t_1, t_2, \ldots t_n]$ and the probability distribution representation of the topic $C = [c_1, c_2, \ldots c_m]$.

For news A and B, their headline similarity can be directly defined by the cosine similarity of the vector:

$$titlesim(title_A, title_B) = \frac{1}{2}cossim(CLM(title_A), CLM(title_B)) + \frac{1}{2} \tag{13}$$

Through the topic similarity matrix $TM_{2k \times 2k}$, the topic similarity $TM_{ij}$ between any two topics i and j can be obtained, the probability of news A belonging to topic i is $c_{Ai}$, the probability of news B belonging to topic j is $c_{Bj}$, so the topics of news AB are similar. The degree is:

$$topicsim(topic_A, topic_B) = \sum_{i=1}^{m}\sum_{j=1}^{m} c_{Ai} c_{Bj} TM_{ij} \tag{14}$$

Therefore, the similarity of news AB is:

$$newssim(A,B) = \alpha \cdot titlesim(title_A, title_B) + \beta \cdot topicsim(topic_A, topic_B) \tag{15}$$

where α and β are the weight coefficients of title similarity and topic similarity, respectively.

3.2.3. Based on improved Single-Pass Chinese-English bilingual news clustering

The algorithm flow of applying the traditional Single-Pass to the bilingual news clustering scene in this paper is as follows:

1. There is a news text sequence $n_1, n_2, \ldots$, specify the news similarity threshold news_threshold;
2. For news $n_i$, calculate Its similarity with all clusters, find the cluster with the largest similarity, if the maximum similarity is greater than news_threshold, add the news $n_i$ to this cluster; otherwise, the news $n_i$ is clustered independently.

The similarity between the sample and the cluster in the original Single-Pass is the sample Maximum similarity to all news in this cluster. It can be seen from the above process that the cluster attribution of the current sample can be found by traversing all samples at most once; if the current total number of samples is N, the time complexity is O (N), which is very efficient compared with the complexity of other non-incremental clustering algorithm $O(n^2)$.

Although the advantages of Single-Pass applied to news clustering are obvious, the shortcomings should not be ignored:

1. The time complexity of O(N) is still too large, especially when performing topic-level coarse-grained clustering, where news is clustered in a few;
2. When performing event-level fine-grained clustering, the temporal characteristics of streaming data are not reflected;
3. The accuracy of Single-Pass is not high, because the cluster of samples belongs. After it is determined, it will not change, even if the density of the two clusters is reachable, the number of clusters will only increase, and there is no cluster merging process.

For the original Single-Pass directly applied to the three news clustering scenarios Defects, this paper makes the following improvements to Single-Pass:

1. Define the mean value of news in the cluster (the mean of the headline vector and the mean of the topic probability distribution) as the cluster center, and directly calculate the distance between the news and the cluster;
2. In the event When clustering high-level fine-grained news, the news release time parameter is added. In highly similar news, the time is more likely to be the same event;
3. Specify the cluster merging threshold. As news is added continuously, the cluster center is constantly changing. Each clustering, if the distance between clusters is less than the cluster merging threshold, the two clusters will be merged [37].

If the existing sample point is m, there is a cluster $N = \{n_1, n_2, \ldots n_k\}$, then in the original Single-Pass algorithm, the distance between m and C is defined as:

$$distance(m, N) = max(distance(m, n_1), distance(m, n_2), \ldots distance(m, n_k)) \quad (15)$$

A piece of news can be defined by the title vector plus the topic probability distribution, title vector $T = [t_1, t_2, \ldots t_{768}]$, topic distribution $C = [c_1, c_2 \ldots c_{22}]$, the average of the headline vectors of all news in the cluster, and the average of the topic distributions of all news can be defined as the following cluster center:

$$N_{avg} = T_{avg} + C_{avg} \quad (16)$$

The cluster center can be regarded as a virtual representative news, so the distance between news and the cluster can be calculated directly according to the news similarity formula. If the number of current clusters is k, the time complexity is O(k), which can be achieve the effect of real-time clustering.

When a cluster is represented by the cluster center, the news similarity formula can be applied to calculate the distance between clusters [38]. Specify the similarity threshold between clusters and cluster threshold. After one clustering is completed, if the similarity between the two clusters is greater than cluster threshold, the two clusters will be merged and the cluster center will be recalculated.

The process of the coarse-grained news clustering algorithm at the topic level is as follows:

1. Calculate the similarity between the newly added sample a and the center of each cluster, and take the cluster c;
2. If a and c is less than the threshold news threshold, sample a Independent clustering, otherwise, add sample a to cluster c, recalculate and persist the cluster center of c;
3. If c changes, calculate c and other clusters, and take the similarity cluster t, if t and c is greater than the threshold cluster threshold, merge c and t, recalculate and persist the cluster center of the merged cluster.

When doing fine-grained news clustering at the event level, we can increase the title's weight of similarity $\alpha$, reduces the weight of topic similarity, because the title is often a high summary of news events [39]. Even so, in many similar events, the headlines still cannot distinguish, for example, there are two NBA sports news headlines are "Grizzlies beat the Nets", but these are events of different time periods. Therefore, in the event-level clustering, the input order of news is arranged in the order of release time, adding the time interval parameter $\Delta d$, and specifying the news time threshold time_threshold, for any two news in the cluster, if the time interval $\Delta d >$ time_threshold, then split into two sub-clusters.

The algorithm flow of event-level fine-grained news clustering is as follows:

1. Calculate the similarity between the new sample a and the center of each cluster, and arrange the clusters $c_1, c_2, c_3, ……c_k$ in descending order of similarity;
2. If the similarity between a and $c_i$ is greater than the threshold news_ Threshold and the time interval $\Delta d$ of any news in a and $c_i$ is less than the time threshold time_ threshold, add sample a to cluster c, recalculate and persist c the cluster center Otherwise i+1 repeat step2. If i = k, a clustered independently;
3. If c changes, calculate c and other clusters, and take the cluster t, if t and c is greater than the threshold cluster_threshold, merge c and t, recalculate and persist the cluster center of the merged cluster.

## 4. Experiments

### 4.1. Results of Coarse-grained clustering

At the topic level When clustering news at the topic level, if two news stories are both military topics (regardless of whether they talk about the same event or not), we consider the two news stories to belong to the same cluster. Therefore, it is necessary to increase the topic weight $\alpha$ and reduce the title weight $\beta$.

Parameter values are shown in Table 2:

**Table 2.** Parameter values.

| $\alpha$ | $\beta$ | news_threshold | cluster_threshold |
|---|---|---|---|
| 0.25 | 0.75 | 0.7 | 0.82 |

The results without inter cluster clustering are shown in Table 3:

**Table 3.** Results without inter cluster clustering.

| Cluster Number/Label | Military | Politics | Economy | Entertainment | Sports | Science | Automobile | Total |
|---|---|---|---|---|---|---|---|---|
| 1 | 2 | 1 | 64 | 3 | 2 | 1 | 0 | 73 |
| 2 | 2 | 12 | 44 | 0 | 5 | 0 | 0 | 63 |
| 3 | 0 | 2 | 0 | 0 | 16 | 6 | 32 | 56 |
| 4 | 7 | 4 | 0 | 0 | 7 | 3 | 170 | 191 |
| 5 | 9 | 3 | 6 | 0 | 1 | 6 | 65 | 90 |
| 6 | 2 | 4 | 18 | 134 | 33 | 2 | 6 | 199 |
| 7 | 6 | 3 | 2 | 0 | 0 | 1 | 4 | 16 |
| 8 | 5 | 19 | 11 | 10 | 23 | 11 | 1 | 80 |
| 9 | 7 | 2 | 14 | 3 | 156 | 4 | 13 | 199 |
| 10 | 17 | 98 | 8 | 3 | 2 | 3 | 0 | 131 |
| 11 | 14 | 113 | 12 | 1 | 0 | 4 | 0 | 144 |

| 12 | 7 | 0 | 12 | 2 | 15 | 102 | 3 | 141 |
| 13 | 5 | 1 | 9 | 0 | 3 | 112 | 3 | 133 |
| 14 | 19 | 4 | 2 | 143 | 13 | 0 | 1 | 182 |
| 15 | 80 | 14 | 8 | 0 | 1 | 31 | 2 | 136 |
| 16 | 117 | 0 | 5 | 0 | 7 | 13 | 0 | 142 |
| 17 | 1 | 20 | 85 | 1 | 16 | 1 | 0 | 124 |

n be seen that there are many clusters of noise data, such as clusters 2, 3, 5, 7, and 8. This is because the distance between the two clusters is long at the beginning, but with the addition of sample points, the center of the cluster changes continuously so that the two the distance between them is getting closer and closer and should be merged into the same cluster.

The results of adding inter-cluster clustering are shown in Table 4:

Table 4. Results of adding inter-cluster clustering.

| Cluster Number/Label | Military | Politics | Economy | Entertainment | Sports | Science | Automobile | Total |
|---|---|---|---|---|---|---|---|---|
| 1 | 4 | 13 | 108 | 3 | 7 | 1 | 0 | 136 |
| 2 | 16 | 9 | 6 | 0 | 24 | 15 | 267 | 337 |
| 3 | 2 | 4 | 18 | 134 | 33 | 2 | 6 | 199 |
| 4 | 6 | 3 | 2 | 0 | 0 | 1 | 4 | 16 |
| 5 | 5 | 19 | 11 | 10 | 23 | 11 | 1 | 80 |
| 6 | 7 | 2 | 14 | 3 | 156 | 4 | 13 | 199 |
| 7 | 31 | 211 | 20 | 4 | 2 | 7 | 0 | 275 |
| 8 | 12 | 1 | 21 | 2 | 18 | 214 | 6 | 274 |
| 9 | 19 | _ | 2 | 143 | 13 | 0 | 1 | 182 |
| 10 | 197 | 14 | 13 | 0 | 8 | 44 | 2 | 278 |
| 11 | 1 | 20 | 85 | 1 | 16 | 1 | 0 | 124 |

t after joining between clusters, similar clusters are merged into the same cluster, and the number of clusters is significantly reduced. Because of the number of clusters is more than the number of labels, and the prediction category cannot be directly seen like the subject headings of LDA, the Kappa coefficient method cannot be used to evaluate the clustering results.  We use precision to evaluate clustering results (recall is meaningless in this scenario, because news under the same label is internally differentiated, discussed above).

From Table 3 and Table 4, it can be seen that the number of news in clusters 4 and 5 accounts for a very small proportion, accounting for 0.76% and 3.8% of the total number of news respectively, which are noise points, so they are not included in the statistics. The maximum number of news under each label of each cluster is used as the accurate clustering result, and the results are shown in Table 5:

Table 5. Accurate clustering results.

| Cluster Number | Accuracy Rate |
|---|---|
| 1 | 0.7941 |
| 2 | 0.7923 |
| 3 | 0.6734 |
| 6 | 0.7839 |
| 7 | 0.7673 |
| 8 | 0.7810 |
| 9 | 0.7857 |
| 10 | 0.7164 |
| 11 | 0.6855 |
| AVG | 0.7560 |

In order to examine the effect of the model on bilingualism, the statistics of the number of languages in the cluster news are shown in Table6:

**Table 6.** Accurate clustering results.

| Cluster number | Main category Chinese news | Main category English news | Main category Total news | Chinese proportion |
|---|---|---|---|---|
| 1 | 55 | 53 | 108 | 0.5093 |
| 2 | 140 | 127 | 267 | 0.5243 |
| 3 | 72 | 62 | 134 | 0.5373 |
| 6 | 69 | 87 | 156 | 0.4423 |
| 7 | 108 | 103 | 211 | 0.5118 |
| 8 | 107 | 107 | 214 | 0.5000 |
| 9 | 68 | 75 | 143 | 0.4755 |
| 10 | 103 | 94 | 197 | 0.5228 |
| 11 | 40 | 45 | 85 | 0.4706 |
| total | 762 | 753 | 1515 | 0.5030 |

The distribution of Chinese and English news of the main category is very close to1:1, and the bilingual effect of clustering is exactly good.

*4.2. Event-level fine-grained clustering*

In event-level fine-grained clustering, even if two news articles belong to the same topic, if they describe two news events, they should still be distributed in two clusters in the clustering result [40]. Generally, news headlines are general descriptions of news events, so the weight parameter $\alpha$ of the headlines should be increased during clustering. The parameters are shown in Table7:

**Table 7.** Parameter values.

| $\alpha$ | $\beta$ | news_threshold | cluster_threshold | time_threshold |
|---|---|---|---|---|
| 0.9 | 0.1 | 0.7 | 0.8 | 365(days) |

The experimental data are composed of sports news, including "related to the opening ceremony of the Winter Olympics", "related to the NBA finals in 2021" and "related to the World Cup". The quantity composition is shown in Table 8:

**Table 8.** The quantity composition.

| News events | Number of news (Chinese -English) |
|---|---|
| Opening Ceremony of the Pingchang Winter Olympic Games | 20-20 |
| Opening Ceremony of the Beijing Winter Olympics Games | 20-20 |
| The NBA Finals | 40-40 |
| The World Cup in Brazil | 20-20 |
| The World Cup in Russia | 20-20 |

The clustering results without the time parameter time_threshold are shown in Table 9:

**Table 9.** The clustering results without the time parameter.

| Cluster number\Event | Pyeongchang Winter Olympics Opening Ceremony | Beijing Winter Olympics Opening Ceremony | NBA Finals | Brazil World Cup | Russia World Cup | Total |
|---|---|---|---|---|---|---|
| 1 | 33 | 36 | 11 | 4 | 0 | 84 |
| 2 | 2 | 0 | 59 | 0 | 0 | 61 |
| 3 | 4 | 2 | 5 | 29 | 37 | 77 |
| 4 | 1 | 2 | 5 | 7 | 3 | 18 |

The text continues here (Figure 2 and Table 2). It can be seen that cluster 4 is the noise data. Only by observing cluster 123, it is found that the news related to the PyeongChang Winter Olympics and the opening ceremony of the Beijing Winter Olympics is not clustered independently, and the news of the World Cup in Brazil and the World Cup in Russia are also mixed. in cluster 3.

After adding the time parameter time_threshold=365, the clustering results are shown in Table 10:

**Table 10.** The clustering results with the time parameter.

| Cluster number\Event | Pyeongchang Winter Olympics Opening Ceremony | Beijing Winter Olympics Opening Ceremony | NBA Finals | Brazil World Cup | Russia World Cup | Total |
|---|---|---|---|---|---|---|
| 1 | 31 | 4 | 5 | 2 | 0 | 42 |
| 2 | 2 | 32 | 6 | 2 | 0 | 42 |
| 3 | 2 | 0 | 59 | 0 | 0 | 61 |
| 4 | 0 | 0 | 3 | 25 | 4 | 32 |
| 5 | 4 | 2 | 2 | 4 | 33 | 45 |
| 6 | 1 | 2 | 5 | 7 | 3 | 18 |

The text continues here (Figure 2 and Table 2). It shows that after adding time_threshold, similar events are clearly distinguished.

The precision, recall, and F-value of the clustering are shown in Table 11:

**Table 11.** The precision, recall, and F-value of the clustering.

| News event | precision | recall | F1 value |
|---|---|---|---|
| Opening Ceremony of the Pingchang Winter Olympic Games | 0.7381 | 0.7750 | 0.7561 |
| Opening Ceremony of the Beijing Winter Olympics Games | 0.7619 | 0.8000 | 0.7805 |
| The NBA Finals | 0.9672 | 0.7375 | 0.8369 |
| The World Cup in Brazil | 0.7813 | 0.6250 | 0.6945 |
| The World Cup in Russia | 0.7333 | 0.8250 | 0.7765 |

Overall accuracy is 0.75, kappa coefficient is 0.7069.

## 5. Conclusions

In this study, we have constructed an efficient cross-lingual news automatic classification model that does not rely on translation systems or human annotation, achieving true cross-lingual clustering. However, there are still some limitations and shortcomings: regarding event-level fine-grained news clustering, we assume that the title should summarize the news event, and we mainly judge whether they describe the same event through the similarity of title vectors. However, some news writers may not follow this approach when creating headlines. Another limitation is that since we are dealing with streaming text data, subsequent news is unknown, and the model parameters selected may not be suitable for later input samples. The inability to dynamically adjust parameters once they are set is another shortcoming of this model. Additionally, during fine-grained clustering under big data conditions, a large number of clusters may lead to extended single clustering times.

In the experiments, we validated that the proposed approach can indeed solve the cross-lingual news clustering problem in both Chinese and English through evaluation metrics such as accuracy, recall, F1 score, and Kappa coefficient. For topic-level coarse-grained news clustering, the accuracy reached 75.6%, and for event-level fine-grained news clustering, the accuracy reached 75%, with a Kappa coefficient of 0.7069.

It is worth mentioning that our proposed method is able to scale in large scale networks, and our future work will mainly focus on extending it in federated learning scenarios [41][42].

## 6. Patents

**Funding:** This work is supported by the science and technology project of State Grid Corporation of China (The key technologies and applications of brand influence evalution and monitoring fits into the layout of "One body and four wings",grant no.1400-202257240A-1-1-ZN).

**Data Availability Statement:** Not applicable.

**Conflicts of Interest:** The authors declare no conflict of interest.